# A Novel Traffic Simulation Framework for Testing Autonomous Vehicles Using SUMO and CARLA

Pei Li, Arpan Kusari, David J. LeBlanc

University of Michigan Transportation Research Institute

## Abstract

Traffic simulation is an efficient and cost-effective way to test Autonomous Vehicles (AVs) in a complex and dynamic environment. Numerous studies have been conducted for AV evaluation using traffic simulation over the past decades. However, the current simulation environments fall behind on two fronts– (1) the background vehicles (BVs) fail to simulate naturalistic driving behavior and (2) the existing environments do not test the entire pipeline in a modular fashion. This study aims to propose a simulation framework that creates a complex and naturalistic traffic environment. Specifically, we combine a modified version of the Simulation of Urban MObility (SUMO) simulator with the Cars Learning to Act (CARLA) simulator to generate a simulation environment that could emulate the complexities of the external environment while providing realistic sensor outputs to the AV pipeline. In a past research work, we created an open-source Python package called SUMO-Gym which generates a realistic road network and naturalistic traffic through SUMO and combines that with OpenAI Gym to provide ease of use for the end user. We propose to extend our developed software by adding CARLA, which in turn will enrich the perception of the ego vehicle by providing realistic sensors' outputs of the AVs' surrounding environment. Using the proposed framework, AVs' perception, planning, and control could be tested in a complex and realistic driving environment. The performance of the proposed framework in constructing output generation and AV evaluations are demonstrated using several case studies.

## Introduction

With the advent of autonomous vehicles (AVs), there has been an increased focus on rigorous testing and validation under significant amount of environmental and dynamic uncertainty. Especially, there is a huge divide between autonomously driving some demonstration vehicles being managed by professional safety drivers under carefully chosen road and weather conditions and releasing a large volume of production ready vehicles that would need to encounter different variabilities (often outside the scope of the original testing) in the real world. Overcoming this gap is made harder without complete knowledge of the variability in the driving environment.

One possible facet of solving the testing and validation problem is by utilizing simulation to provide repeatable scenarios under different dynamic and environmental conditions. Large scale simulation based testing is a cost effective and efficient approach which can provide the first step towards testing highly complex systems such as AVs. There have been various open-source and commercial simulation software proposed for AV testing such as CARLA, Metamoto, Airsim etc. These simulation software contain highly detailed vehicular traffic models, high fidelity simulation and generating testing scenarios. However, they lack in two key aspects:

- The background vehicles (BVs) behave in a pre-determined manner which is completely antithetical to real-world variability. This leads to BVs being preemptively cautious in most cases which does not provide adequate testing for AVs.
- There is a need to test the AV pipeline in a divide-and-conquer fashion where instead of testing the system as a whole, we test individual sub-systems to understand the error characteristics and robustness. This can provide the AV developers with a lot of flexibility in terms of only testing sub-systems which have been updated and understanding (and quickly correcting) the failure of an individual sub-system. Another advantage of such an approach is that the error budget of the entire system is the summation of error budget of the individual sub-systems; figuring out the error budget of the sub-systems can lead to efficiency in developing a production ready AV.

We tackle both these problems in this paper by combining CARLA with a modified version of SUMO. In our past work [1], we combined the SUMO simulator with OpenAI Gym to provide convenience and accessibility to users. Using naturalistic data, we estimated the parameters of IDM model which we then sampled to create realistic BVs. In this work, we combine the earlier work with CARLA to provide high fidelity perception by outputting different kinds of sensor outputs. The sensor outputs can then be fed into the perception module of the AV pipeline. The output from perception module can be judged against the simulation "ground-truth". The features of the neighboring vehicles can be output for the decision making and control modules to navigate the vehicle. We show that we can utilize our simulation system, henceforth known as CARLA-SUMO-Gym, to run high fidelity simulations without any prior knowhow and with a handful of Python commands.

Finally, we showcase benchmarking studies using CARLA-SUMO-Gym to compare and contrast the performance of open-source deep neural network (DNN) based object detectors in various different environmental conditions. There has been very little work done in combining gaming engines with SUMO [2, 3] and from our perfunctory search, they do not focus on the modular testing of AVs.

Our paper is set up as follows: in the next section, we provide an overview of our simulation platform; in the following section, we discuss the different DNN based object detectors and metrics for performing comparisons between the detections and ground truth and finally, we present the results of using the object detectors for detecting vehicles in different weather conditions and discussions.

## Overview of CARLA-SUMO-Gym

We aim to provide users maximum convenience and accessibility by abstracting the process of scenario generation and high-fidelity simulation without the burden of creating these simulations from scratch. In that regard, we combine our previous package, SUMO-Gym with CARLA to provide these benefits for AV developers without the additional cost. In the following subsections, we explain the different constituents of our simulation package and the combined package.

### *OpenAI Gym*

Gym is an open source Python library created for testing reinforcement learning algorithms in various simulation environments [4]. A key feature of the library is ease-of-use for the user due to its standard API and a common interface for all simulation environments. This then provides a way of abstracting the process of loading a particular simulation environment and controlling the agents in the simulation environment through very simple, intuitive commands without having explicit knowledge of the inner workings of the environment.

### *Simulation of Urban Mobility (SUMO)*

SUMO has been developed at the German Aerospace Center as a microscopic simulator to model traffic flow and predict traffic using simulations [5, 6]. SUMO can be used to simulate traffic flow in large scale setups without sacrificing on speed. It can import road network from various different simulators or open-source maps, provides tools for easy demand generation and incorporates different realistic longitudinal and lateral vehicle models. Each vehicle in the simulation can be independently configured with different departure and arrival properties. Any vehicle can also be controlled via an external application using an API called Traffic Control Interface (TraCI) [7] . All these features make SUMO incredibly light weight but robust for simulating repeatable scenarios.

However, SUMO suffers from some limitations which make it difficult to be used as a standalone testing and validation platform for AVs. Firstly, it is very cumbersome to set up a road network, create the demand network and then run a simulation designating one particular vehicle as subject vehicle. Secondly, providing an action in the road aligned coordinate frame (as an output from the controls module) is not straightforward. Finally, it cannot be used to evaluate the perception sub-system of AV pipeline at all since it has very low fidelity in terms of visualization.

### *Cars Learning to Act (CARLA)*

CARLA simulator was developed as a collaboration of Intel Labs, Toyota Research Institute and Computer Vision Center, Barcelona. It has been built as an open-source layer over Unreal Engine 4 with the goal of flexibility and realism in high fidelity simulations. CARLA includes urban layouts and a variety of models in different classes such as vehicles, pedestrians, buildings and traffic signs. A very important feature of the simulator is its inbuilt sensor suite setup which can be placed on any vehicle such as camera, LiDAR, radar etc. Another way of adding variability to the simulated scenarios is through simulation of different weather conditions which can be done easily using CARLA.

While CARLA is adept at producing high fidelity simulations, generation and movement of traffic is not its strong suite. We next present our simulator approach which combines the strength of these three packages into one simulation package.

### *CARLA-SUMO-Gym*

The simulation package is detailed in Figure 1. We choose two classes of scenarios: highway and urban and provide some example road network files. Given the network files, we also extract the routes in the network. We then estimate data distributions from naturalistic driving data (NDD) and sample the background vehicles in each given route with the values from the distributions. The network and the demand files are input as configuration files to the SUMO simulator which runs the simulation. We create a Python package, inherited from the environment class of Gym, to control the ego vehicle, provide the observations and advance at each step given the action from the AV pipeline. We provide two different kinds of observations: using CARLA, we provide the raw sensor outputs (Camera, LiDAR and radar) and we provide high level features (positions and speeds in road aligned coordinate frame) of the neighboring vehicles. We take the longitudinal and lateral accelerations from the AV pipeline as the input at each step.

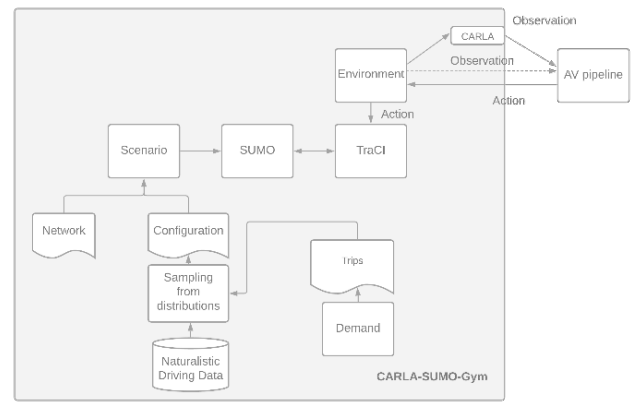

Figure 1 Flowchart of different components of CARLA-SUMO-Gym

We show the simultaneous visualization in CARLA and SUMO in Figure 2. From the visualizations, we can immediately see the differences between fidelity of the two simulators even if they generate the same scene. In Figure 3, we show the example camera images and LiDAR point clouds that are passed to the user as observations. The camera captures an RGB image of image size 800x600 pixels from the vehicle's point-of-view at a given sensor frequency. In order to provide realistic images, a set of post-process effects can be applied such as lens flares and grain jitter which can help in understanding the effects of these on object detections. In addition, different types of distortions can also be applied to these camera images representing miscalibrations. For LiDAR simulations, CARLA replicates a rotating LiDAR based on ray-casting. Essentially, the points are computed in the vertical direction spaced by a vertical field-of-view (FOV). The rotation of the sensor is computed using a horizontal step angle. The intensity of each LiDAR point is also calculated using the following formula: $\frac{I}{I_0} = e^{-a \cdot d}$, where $I_0$ is the actual intensity, $a$ is the attenuation coefficient and $d$ is the distance of the point from the sensor.

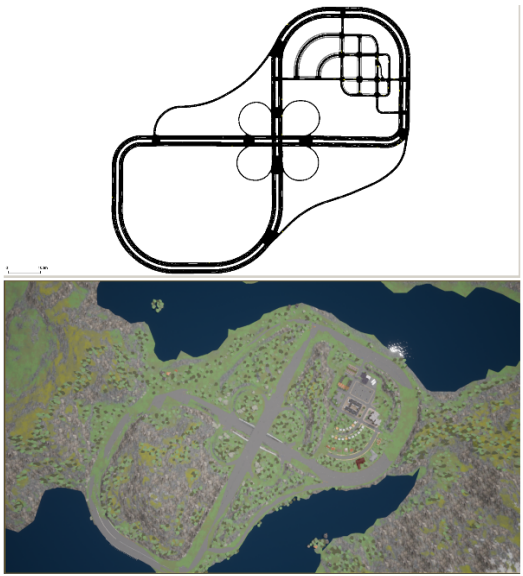

Figure 2 An example visualization in CARLA and SUMO

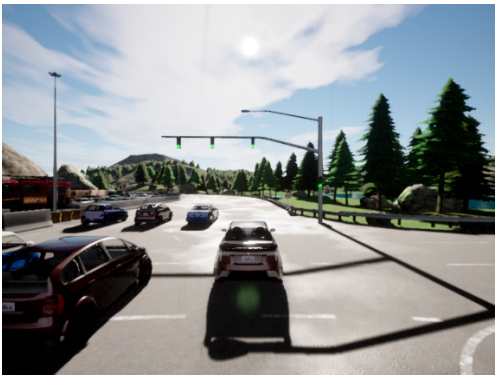

Figure 3 Camera image and LiDAR point cloud generated using CARLA

## Deep Neural Network-based Object Detection

The benchmark model used for object detection is provided by Facebook's detectron2 [8]. Detectron2 contains a variety of state-of-the-art models, which could be used for all kinds of tasks of image processing (Figure 4), such as object detection, keypoint detection, semantic segmentation, etc. Detectron2 provides a convenient way for its users to use different models without training them from scratch. The benchmark model used in this study is a Mask R-CNN (Regional Convolutional Neural Network) which has a ResNet-50 [9] and Feature Pyramid Network (FPN) as a backbone with standard convolutional neural network and fully connected heads for mask and box prediction [8]. The model was pretrained on the COCO dataset [10], which contains 330K images and 80 object categories. Since the focus of this paper is to illustrate the feasibility of using CARLA-SUMO-Gym for testing autonomous vehicles. The object detection in this paper aims to detect the surrounding vehicles of the ego-vehicle, using images captured by the camera installed in the front of the ego-vehicle. Other objects such as pedestrians, traffic lights, and traffic signs could also be detected but are not investigated in this paper.

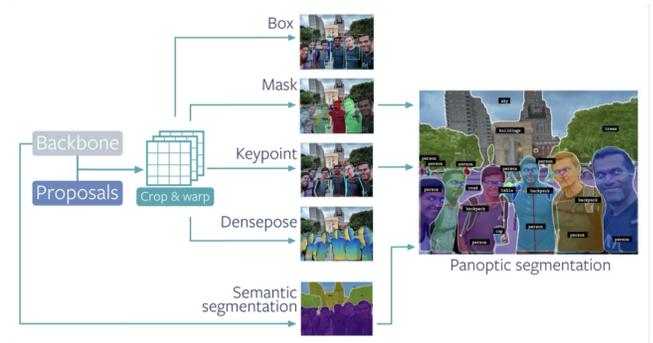

Figure 4 Outputs from detectron2 [8]

## Experimental Design and Results

A case study was conducted to compare the performance of the benchmark object detection model on different weather conditions. Two scenarios were simulated using CARLA, the scenario of the morning (Figure 5 (a)) and the night (Figure 5 (b)). Specifically, the scenario of the morning has perfect lighting condition, while the scenario of the night has poor lighting condition. A camera was placed in the front of the ego-vehicle to capture images for object detection. The background vehicles were generated through SUMO. Figure 6 shows the road network used in this paper. Three experiments were conducted for each scenario, while each experiment contained 200 vehicles. A random vehicle was selected as the ego-vehicle, which was equipped with a camera to capture images of the surrounding environment.

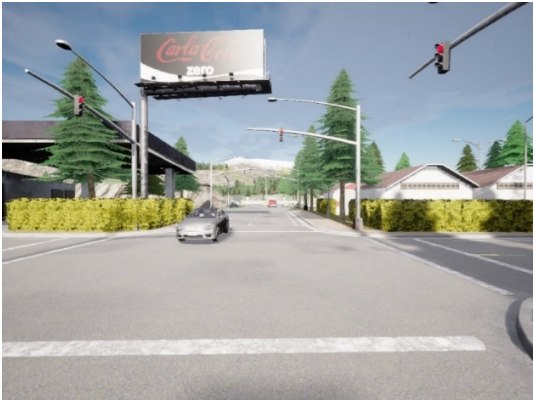

(a) The scenario of the morning

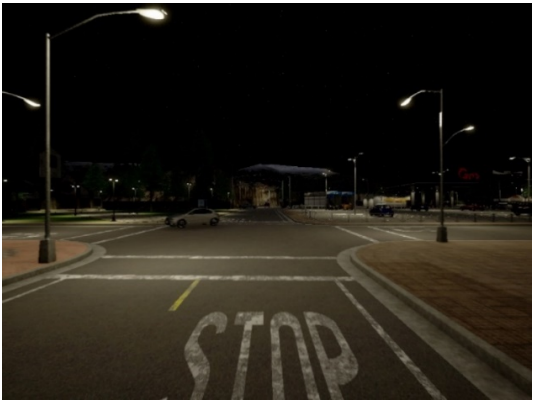

(b) The scenario of the night

Figure 5 Examples of two scenarios

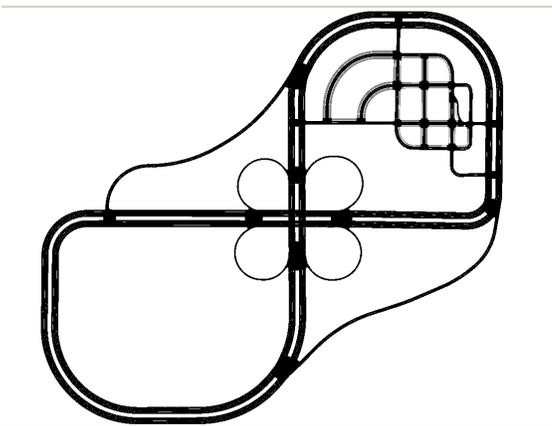

Figure 6 The road network

Intersection over Union (IoU) was used to measure the performance of the object detection. IoU was estimated by dividing the area in common between two bounding boxes by the area included in at least one of the boxes (Figure 7) [11]. To estimate IoUs of an object detection model, bounding boxes of ground truth should be provided. Traditional ways of obtaining ground truth usually require manual labeling, which is time-consuming and lacks efficiency. CARLA provides a convenient way to obtain the ground truth for object detection. The bounding boxes of surrounding vehicles could be automatically extracted through the Python API provided by CARLA (Figure 8).

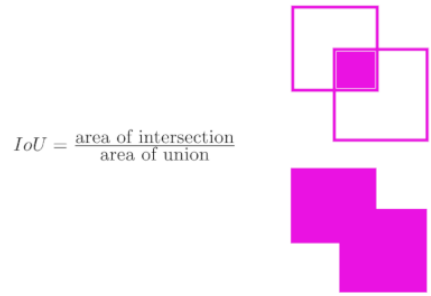

Figure 7 The illustration of IoU

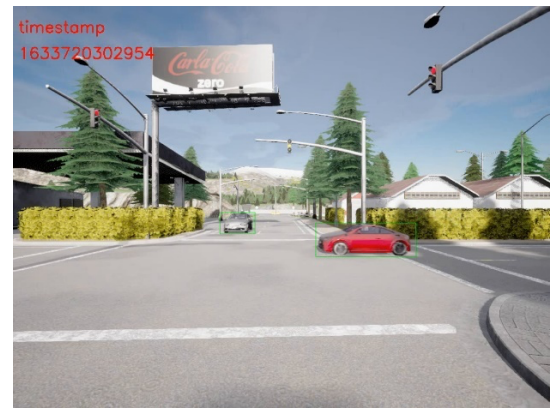

Figure 8 An example of ground truth generated by CARLA

Two metrics, precision and recall were used to evaluate the performance of the object detection model. Precision and recall were estimated using equations (1) and (2). True Positive (TP) and False Positive (FP) were estimated based on the selected IoU threshold. Specifically, if the IoU threshold is selected as 0.5, the object detection result will be determined as a TP if its IoU is greater than or equal to 0.5. While a FP represents that the IoU is smaller than 0.5. Figure 9 shows an example of different IoUs. If we select the IoU threshold as 0.5. The value of TP will be 3 and the value of FP will be 1. If we select the IoU threshold as 0.7. The value of TP will be 2 and the value of FP will be 2. Moreover, A False Negative (FN) means that the object detection model fails to detect the object, which means the model does not detect vehicles in this paper.

$$Precision = \frac{TP}{TP + FP} \quad (1)$$

$$Recall = \frac{TP}{TP + FN} \quad (2)$$

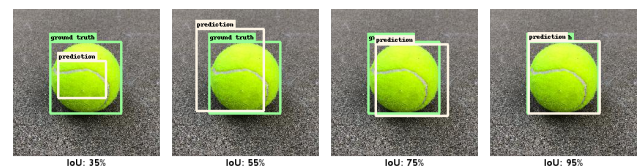

Figure 9 An example of different values of IoUs [12]

Figure 10 shows an example of the object detection model. The outputs from the model are the coordinates of the bounding boxes, which can then be used to estimate the values of IoU, precision, and recall. Table 1 shows the experimental results under different IoU thresholds ranging from 0.5 to 0.8. The object detection model achieved better results on the morning scenario than the night scenario with higher values of precision and recall. The results were as expected since the night scenario had worse visibility compared to the morning scenario. Moreover, the object detection model had much higher values of recall on the morning scenario than the night scenario. However, the model had slightly higher values of precision on the morning scenario than the night scenario. The results indicated that the object detection model had decent performance on both scenarios in terms of correctly detecting vehicles from all detected results. Nevertheless, the model's ability to correctly detect vehicles from all ground truth vehicles was significantly affected by the weather condition. Moreover, Table 1 shows that the performance of the object detection model is gradually decreasing with the increase of IoU thresholds. The results suggested that using data from one sensor (i.e., camera) may not be enough for detecting objects. Sensor fusion strategies should therefore be applied to improve the perception ability of AVs.

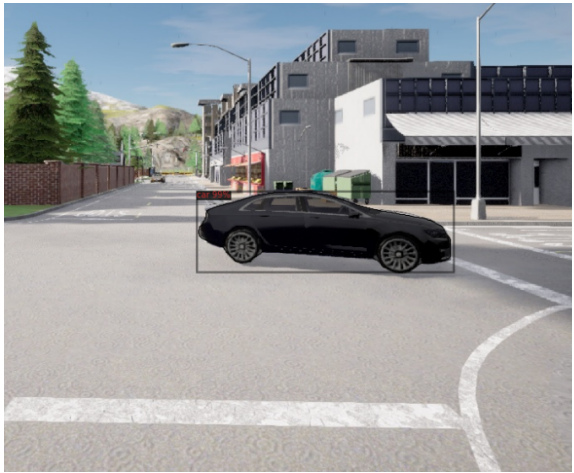

Figure 10 An example of the object detection model

Table 1. Experimental results

|  | Precision | | | | Recall | | | |
| --- | --- | --- | --- | --- | --- | --- | --- | --- |
| **IoU Threshold** | 0.5 | 0.6 | 0.7 | 0.8 | 0.5 | 0.6 | 0.7 | 0.8 |
| **Morning** | 0.82 | 0.81 | 0.79 | 0.47 | 0.41 | 0.40 | 0.40 | 0.28 |
| **Night** | 0.81 | 0.79 | 0.70 | 0.37 | 0.23 | 0.23 | 0.21 | 0.12 |

## Conclusions

This paper aims to propose a framework named CARLA-SUMO-Gym for testing and evaluating AVs. SUMO-Gym was developed in the authors' previous study, which provided a convenient way for micro-level traffic simulation using naturalistic driving data. Users of SUMO-Gym can test their AV pipelines by interacting with SUMO-Gym using the API provided by OpenAI Gym. This paper introduced CARLA as an enhancement for SUMO-Gym. CARLA was used to enrich the traffic simulation by generating a high-fidelity simulation environment. Moreover, the usage of CARLA enabled the testing of AV's perception ability by generating data from various sensors, such as cameras and Lidars. A case study was conducted to evaluate the performance of a benchmark object detection model under different conditions. CARLA was used to simulate two weather conditions, generate outputs from sensors, and generate ground truth for object detection. Experiment results indicated that the benchmark method would have poor performance under the reduced visibility condition, which suggested the necessity of having access to data from other sensors (e.g., Lidars and radars). In the future, the study could be improved by generating outputs from other sensors, as well as providing a variety of scenarios that cover different driving conditions.